\documentclass[a4paper]{article}

\usepackage{INTERSPEECH2022}
\usepackage{color}

\title{Avoid Overfitting User Specific Information in Federated Keyword Spotting}
\name{Xin-Chun Li$^{1}$, Jin-Lin Tang$^{1}$, Shaoming Song$^{2}$, Bingshuai Li$^{2}$, Yinchuan Li$^{2}$, Yunfeng Shao$^{2}$, \\ Le Gan$^{1}$, De-Chuan Zhan$^{1}$ 
\thanks{Supported by National Natural Science Foundation of China (Grant No. 41901270), NSFC-NRF Joint Research Project under Grant 61861146001, and Natural Science Foundation of Jiangsu Province (Grant No. BK20190296). Thanks to Huawei Noah's Ark Lab NetMIND Research Team for funding this research. De-Chuan Zhan is the corresponding author. Email: zhandc@nju.edu.cn}}
\address{
  $^1$State Key Laboratory for Novel Software Technology, Nanjing University\\
  $^2$Huawei Noah’s Ark Lab}
\email{\{lixc, tangjl\}@lamda.nju.edu.cn, ganle@nju.edu.cn, zhandc@nju.edu.cn, \\ \{shaoming.song, libingshuai, liyinchuan, shaoyunfeng\}@huawei.com}

\begin{document}

\maketitle
\begin{abstract}
  Keyword spotting (KWS) aims to discriminate a specific wake-up word from other signals precisely and efficiently for different users. Recent works utilize various deep networks to train KWS models with all users' speech data centralized without considering data privacy. Federated KWS (FedKWS) could serve as a solution without directly sharing users' data. However, the small amount of data, different user habits, and various accents could lead to fatal problems, e.g., overfitting or weight divergence. Hence, we propose several strategies to encourage the model not to overfit user-specific information in FedKWS. Specifically, we first propose an adversarial learning strategy, which updates the downloaded global model against an overfitted local model and explicitly encourages the global model to capture user-invariant information. Furthermore, we propose an adaptive local training strategy, letting clients with more training data and more uniform class distributions undertake more local update steps. Equivalently, this strategy could weaken the negative impacts of those users whose data is less qualified. Our proposed FedKWS-UI could explicitly and implicitly learn user-invariant information in FedKWS. Abundant experimental results on federated Google Speech Commands verify the effectiveness of FedKWS-UI.
\end{abstract}

\section{Introduction}
\label{sec:intro}
Deep learning has been successfully applied to automatic speech recognition (ASR)~\cite{ASR-LAS,ASR-Microsoft}, facilitating the emergence of intelligent voice assistants (e.g., Amazon Alexa). To wake up the smart assistant, some predefined keywords (e.g., ``Alexa") need to be identified precisely from users' speech recordings, i.e., keyword spotting (KWS)~\cite{KWS-DNN,KWS-CNN}. This identification process must be efficient to complete, and the utilized models should have minimal memory footprint. Furthermore, the KWS process should be robust to users with various accents or preferred spoken words.


Recent works utilize various deep networks for KWS~\cite{KWS-CNN,KWS-DSCNN,KWS-MHAttRNN,KWS-ResNet,KWS-Transformer}. These methods take a data centralized training style based on the publicly available benchmark such as Google Speech Commands~\cite{SpeechCommands}. However, there may be significant privacy implications in sharing users' audio recordings, which requires a data decentralized training style for privacy protection. Federated learning (FL)~\cite{FedAvg,FedSurvey-Yang} has been effectively applied for communication efficient decentralized training with basic privacy protection. Although FL could be directly applied to decentralized KWS training, the non-independent and identically distributed data (non-i.i.d. data) poses many challenges~\cite{FedNonIID-Zhao,FedProx}. Non-i.i.d. in KWS refers to the fact that some users only own a small amount of data (i.e., quantity skew), users tend to use different spoken words (i.e., label distribution skew), and users usually have accents (i.e., feature distribution skew). 

This paper investigates FedKWS on Google Speech Commands~\cite{SpeechCommands} with several popular network architectures. Compared with centralized training, we observe a significant performance degradation in FedKWS due to non-i.i.d. data. In fact, the small amount of data and the distribution skew problem make the downloaded global model easily overfit user-specific information. For example, the feature extractor mistakenly takes a user's accent as an important factor, or the classification layer is biased towards a user's commonly spoken words. To solve these challenges and enhance the generalization performance of the federated model, we propose several strategies to avoid the local model overfitting user-specific information.

\section{Related Works}
\label{sec:relate}
Our work is closely related to keyword spotting (KWS)~\cite{KWS-CNN,KWS-DNN} and federated learning (FL)~\cite{FedAvg,FedSurvey-Yang,FedPAN}. Current works formulate KWS as a classification problem, aiming to identify whether a short speech recording is a specific word, silence, or unknown. Considering the success of deep learning, CNN has been applied to KWS~\cite{KWS-CNN}. Depth-separable CNN (DSCNN)~\cite{KWS-DSCNN} is applied to obtain the goal of small footprint memory, and residual network (ResNet)~\cite{KWS-ResNet} is utilized to enhance performances. Recurrent neural networks with multi-head attention (MHAttRNN)~\cite{KWS-MHAttRNN,KWS-Transformer} and varieties of transformers (Transformer)~\cite{KWS-Transformer,KWS-Transformer-AST} have also been applied to KWS and obtain SOTA results. Some other advanced techniques in deep learning have also been verified helpful in KWS~\cite{KWS-SSN}. FL has also been applied to KWS for decentralized training~\cite{FedKWS-Ada,FedKWS-Spec}. \cite{FedKWS-Ada} conducts extensive experiments of FedAvg~\cite{FedAvg} on ``Hey Snips" dataset and uses an adaptive averaging strategy for global model aggregation as done in~\cite{FedOpt}. The work~\cite{FedKWS-Spec} investigates data augmentation and distillation in FedKWS for overcoming resource constraints and example labeling. FL studies have also been presented in ASR~\cite{FedASR-CAFT,FedASR-Diversity,FedASR-LibiriSpeech}. Compared with these studies, we primarily focus on the non-i.i.d. data challenge in FedKWS and propose a novel method to focus on extracting user-invariant information. We investigate our methods with various network architectures and show that our approach is universal.

\section{Background of Federated Learning}
\label{sec:background}
\noindent \textbf{FedAvg}~\cite{FedAvg}: Suppose we have $K$ clients and each client owns a data distribution $\mathcal{D}^k = \mathcal{P}^k(\mathbf{x}, y)$, $k \in [K]$. FL aims to optimize $\min_{\psi}\sum_{k=1}^K p_k \mathcal{L}(\mathcal{D}^k;\psi)$, where $\psi$ denotes the global parameters, $p_k$ denotes the weight of each client. FedAvg~\cite{FedAvg} solves this problem via multiple communication rounds of local and global procedures. During local procedure, a partial set of clients $S_t$ download the global model $\psi_t$ and update it on their local data for multiple steps. During global procedure, the server collects these updated local models (denoted as $\hat{\psi}_t^k$, $k \in S_t$) and aggregates them via parameter averaging, i.e., $\psi_{t+1} \leftarrow \frac{1}{|S_t|}\sum_{k \in S_t} \hat{\psi}_{t}^k$. $t$ denotes the communication round. These two procedures will iterate $T$ rounds until convergence.

\noindent \textbf{Non-I.I.D. Data}: The users' data in FL are often naturally heterogeneous, e.g., the speech data in Google Speech Commands~\cite{SpeechCommands} are collected from users with various accents. As declared in~\cite{FedNonIID-Zhao}, the local update direction will diverge a lot from the global one due to non-i.i.d. data, making the model aggregation inaccurate. FedOpt~\cite{FedOpt} utilizes an adaptive optimization strategy on the server instead of a simple parameter averaging. FedRS~\cite{FedRS} specifies the challenge of label shift across clients and proposes restricted softmax as the solution. FedProx~\cite{FedProx} and FedMMD~\cite{FedMMD} add regularization to prevent local models from being updated too away, which could decrease the weight divergence for better aggregation. Although some FL methods (e.g., FedProx~\cite{FedProx}, FedDyn~\cite{FedDyn}, MOON~\cite{MOON}) could also elaborate a regularization effect during local procedures, they only stay on the parameter or the intermediate feature levels. By contrast, we adversarially update the global model against an overfitted local model and regularize the local procedure on the functional level. Furthermore, we design an adaptive local training procedure from the system scheduling level.

\section{Proposed Methods}
\label{sec:method}
This section proposes two strategies to prevent the global model from overfitting user-specific information (e.g., accents or favorite spoken words) in FedKWS.

\noindent \textbf{\textit{A}dversarial \textit{L}earning against \textit{O}verfitted models (\textit{ALO})}: Clients update the downloaded global model on their data during the local procedure, which could overfit some user-specific information. Specifically, the local data distribution $\mathcal{P}^k(\mathbf{x}, y)$ may diverge significantly from the global data distribution. According to some previous works~\cite{HowTransfer,FedREP,LG-FedAvg}, the lower/higher layers of a neural network tend to be influenced significantly by feature/label distribution skew, i.e., various $\mathcal{P}^k(\mathbf{x})$ or $\mathcal{P}^k(y)$. FedKWS simultaneously faces these two kinds of distribution skew (e.g., accents and favorite spoken words), making the complete model biased towards a specific user during the local procedure. Hence, we must regularize the local training from the functional perspective instead of focusing on specific neural network layers. We resort to private-shared models and adversarially update the global model (shared among users) against overfitted local models (private for each user). Private-shared models are utilized in some recent FL solutions~\cite{FedPHP,FedPS}. Specifically, we build private models $\psi_{\text{p}}^k, k \in [K]$ for each client. We first train private models with the cross-entropy loss $\mathcal{L}(\mathcal{D}^k;\psi_{\text{p}}^k)=E_{\mathbf{x}_i, y_i \sim \mathcal{D}^k }[-\sum_{c=1}^C \mathcal{I}\{y_i=c\}\log [f_{\text{p}}(\mathbf{x}_i)]_c]$, where $\mathcal{I}\{\cdot\}$ is the indicator function and $f_{\text{p}}(\cdot)$ is the prediction function based on private model $\psi_{\text{p}}^k$ that outputs a probability distribution. After abundant training steps, we expect this private model to overfit user-specific data information. Then, we train the global model with the following loss:
\begin{align}
	&\mathcal{L}_{\text{ls}} =E_{\mathbf{x}_i, y_i}\left[-\sum_{c=1}^C [(1-\mu)\mathcal{I}\{y_i=c\} + \mu/C] \log [f(\mathbf{x}_i)]_c\right], \label{eq:ls} \\
	&\mathcal{L}_{\text{adv}}={\color{red} \underbrace{-}_{\text{negative}}}E_{\mathbf{x}_i, y_i }\left[-\sum_{c=1}^C [f_{\text{p}}(\mathbf{x}_i)]_c \log [f(\mathbf{x}_i)]_c\right], \label{eq:adv} \\
	&\mathcal{L}(\mathcal{D}^k;\psi^k)=\mathcal{L}_{\text{ls}}(\mathcal{D}^k;\psi^k) + \lambda \mathcal{L}_{\text{adv}}(\mathcal{D}^k;\psi^k), \label{eq:loss}
\end{align}
where we omit the communication round index $t$ and some other symbols for simplification. $f_{\text{p}}(\cdot)$ represents the function of the overfitted private model while $f(\cdot)$ for the downloaded global model. Eq.~\eqref{eq:adv} could be seen as ``negative distillation", which could push the global model's prediction $f(\mathbf{x}_i)$ away from overfitted areas. Eq.~\eqref{eq:adv} follows the formula of distillation~\cite{Distill,FedDML,GKT} but works significantly different. Label smoothing in Eq.~\eqref{eq:ls} could also regularize the global model not be too over-confident on a specific user's data. We investigate the hyperparameters of $\mu$ and $\lambda$ in ablation studies. 


\begin{figure}[tb]
	\centering
	\centerline{\includegraphics[width=\linewidth]{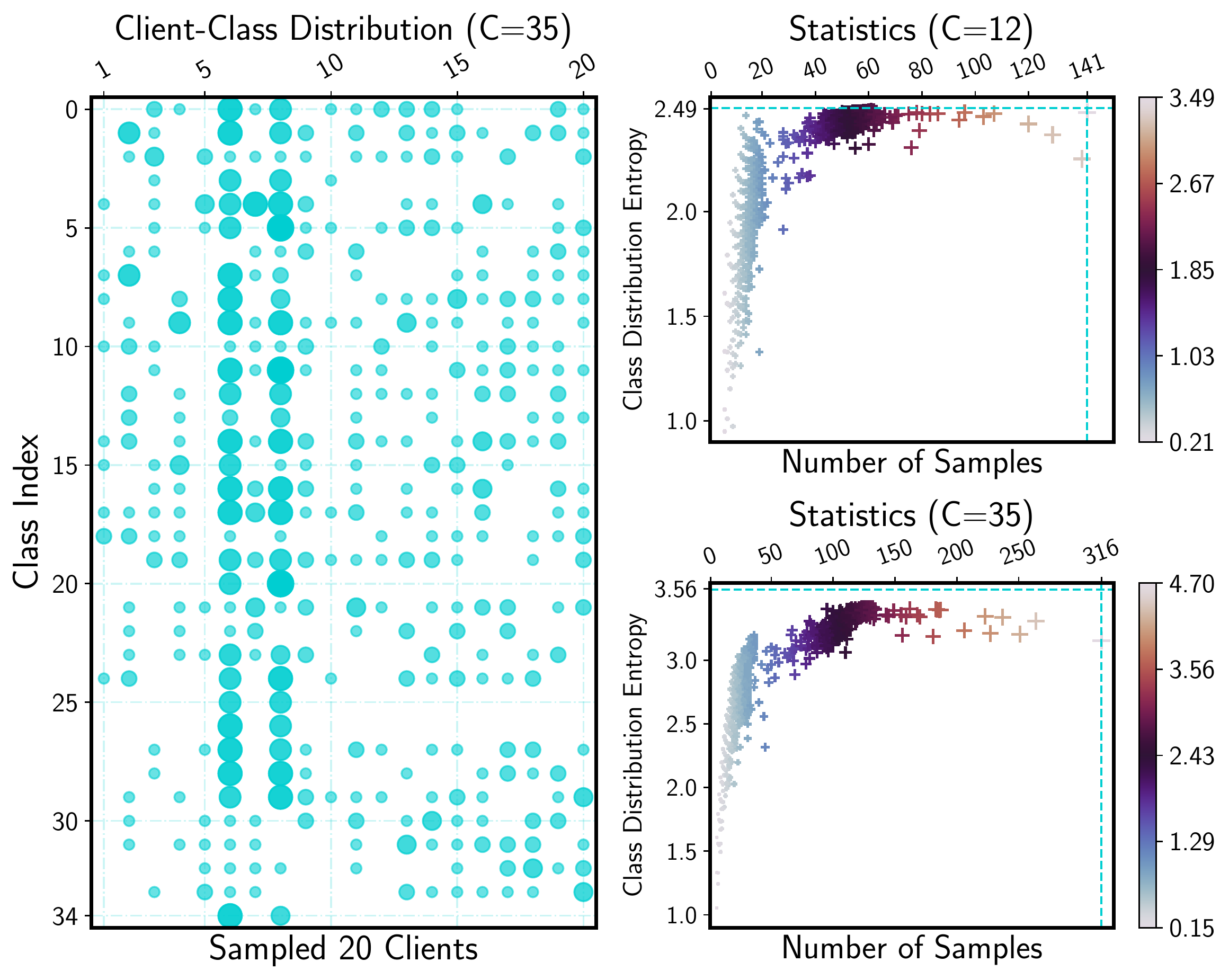}}
	\caption{\textbf{Left}: data heterogeneity in federated Google Speech Commands. We only plot 20 clients (users) in task $35$. \textbf{Right}: number of samples and class distribution entropy of each client (user) in task $12$ and $35$ (each point shows a client).}
	\label{fig:data}
\end{figure}

\noindent \textbf{\textit{A}daptive \textit{L}ocal \textit{T}raining (\textit{ALT})}: Due to data heterogeneity, both amount imbalance and class imbalance could occur in clients' data. The former implies that different clients may own various numbers of training samples. The second one refers to that label distributions may diverge across clients. These two types of imbalance on Google Speech Commands~\cite{SpeechCommands} are shown in Figure~\ref{fig:data}. Intuitively, few training samples could lead to overfitting, and imbalanced data could bias the model towards identifying a user's favorite words. Hence, we encourage clients who own more training data and more uniform class distributions to undertake more local updates.
Formally, in FedAvg~\cite{FedAvg}, every selected client uniformly takes $E$ local training steps without considering the data quality. Assume the $k$th client owns $n_k$ training samples and the class distribution is $\mathbf{q}_k \in \mathcal{R}^C$ with $\sum_{c=1}^C \mathbf{q}_{k,c}=1$ and $\mathbf{q}_{k,c}\geq 0, \forall c$. $C$ is the total number of classes. We calculate the normalized amount of training samples as $\overline{n}_k=n_k/\max_{j=1}^K n_j \in [0, 1]$, and the normalized class entropy as $\overline{e}_k=(-\sum_{c=1}^C \mathbf{q}_{k,c} \log \mathbf{q}_{k,c})/\log C \in [0, 1]$. Then we calculate the harmonic mean of $\overline{n}_k$ and $\overline{e}_k$, i.e., $\overline{r}_k =2\overline{n}_k \overline{e}_k/(\overline{n}_k + \overline{e}_k)$. We use $\overline{r}_k \in [0, 1]$ to measure clients' utility in FL, and we heuristically let clients with larger $\overline{r}_k$ contribute more to FL. That is, we reallocate the computation resources among clients via allowing the $k$th client take on $r_0 * \overline{r}_k * E$ gradient steps, where the determination of $r_0$ should satisfy $\sum_{k=1}^K r_0 * \overline{r}_k * E \approx K * E$ for conservation. Easily, $r_0=K/\sum_{k=1}^K \overline{r}_k$. Although the computation ability of clients should also be considered, we focus on non-i.i.d. data in this work and leave it as future work.
\begin{figure*}[htb]
	\centering
	\centerline{\includegraphics[width=\linewidth]{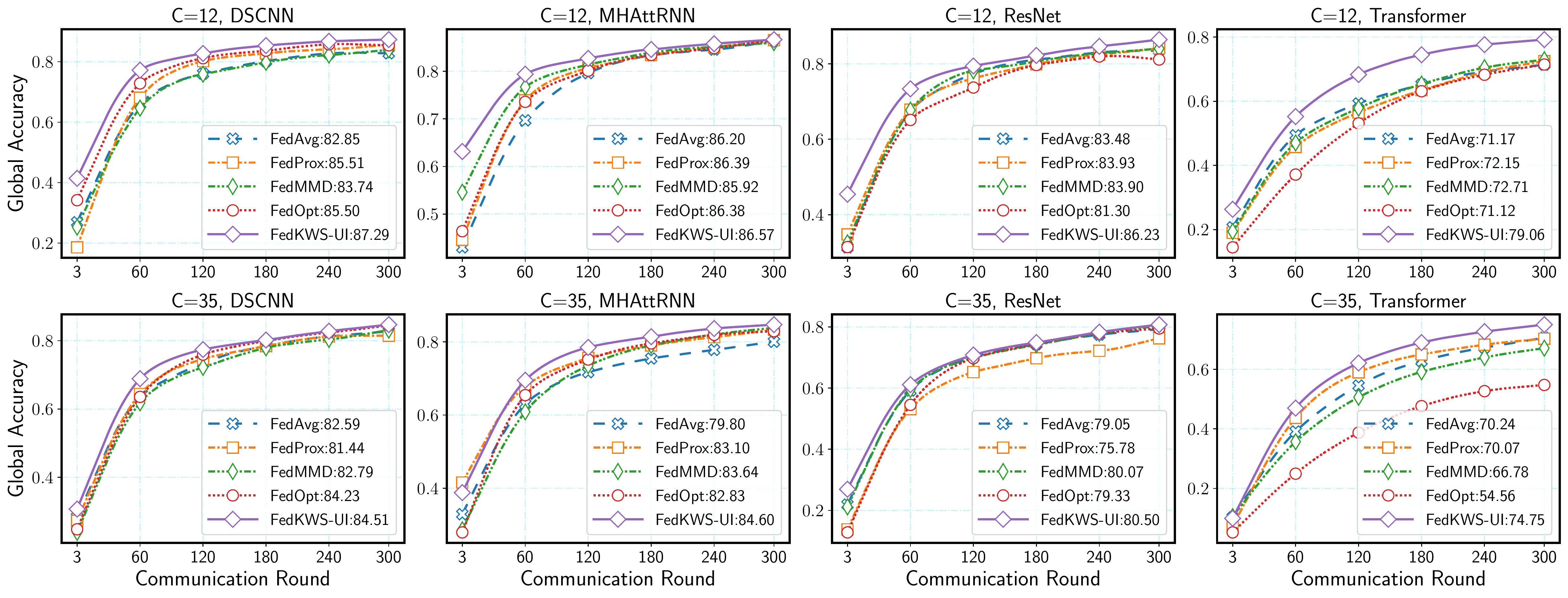}}
	\caption{Comparison results on federated Google Speech Commands. Rows show the results on task 12 and 35, and columns show results of four utilized networks. The legends also show the average accuracy of the final 5 communication rounds.}
	\label{fig:compare}
\end{figure*}

\begin{table}
	\centering
	\caption{\small Detail information of federated Speech Commands.}
	\label{tab:data}
	{
		\begin{tabular}{c|c|c|c|c|c}
			\hline \hline
			$C$ & $K$ & $N$ & Avg.$n_k$ & Max.$n_k$ & M \\
			\hline \hline
			12 & 2,234 & 45.6k & 20.4 & 141 &	4.9k \\ \hline
			35 & 2,434 & 105.5k & 43.3 & 316 & 11.0k \\
			\hline \hline
		\end{tabular}
	}
\end{table}


\section{Experiments}
\noindent \textbf{Datasets}:
We name the proposed method as ``Federated KWS with User-Invariant information" (FedKWS-UI), and investigate it on Google Speech Commands~\cite{SpeechCommands}\footnote{https://pytorch.org/audio/stable/datasets.html} (recommended by FedScale~\cite{FedScale}) to identify whether a 1s-long speech recording is a word, silence, or unknown. The benchmark contains two tasks with 12 classes (10 words, silence, unknown) and 35 classes (35 words). The two tasks contain 2,234 and 2,434 users. We split the data into corresponding clients with each user as a client. The number of total training samples ($N$), the training samples of each client on average (Avg.$n_k$), the number of test samples ($M$) are listed in Table~\ref{tab:data}. The class distributions of randomly selected 20 clients in task $C=35$ are shown in left of Figure~\ref{fig:data}. Larger circles correspond to more samples. The train and test data is split via the provided lists in Google Speech Commands. We extract 40 MFCC features for each 30ms window frame with a stride of 10ms. We also follow the settings in Google Speech Commands: performing random time-shift of $Y \sim [-100, 100]$ milliseconds and adding 0.1 volume background noise with a probability of 0.8.

\begin{table}
	\centering
	\caption{\small Detail of networks and centralized training results.}
	\label{tab:net}
	{
		\begin{tabular}{l|c|c|c|c}
			\hline \hline
			& \multicolumn{2}{c|}{Num.of.Params} & \multicolumn{2}{c}{Centralized Acc.} \\ \hline
			& $C=12$ & $C=35$ & $C=12$ & $C=35$ \\
			\hline \hline
			DSCNN~\cite{KWS-DSCNN} & 169K & 173K & 97.19 & 96.95  \\ \hline
			MHAttRNN~\cite{KWS-MHAttRNN} & 228K & 232K & 97.31 & 97.05 \\ \hline
			ResNet~\cite{KWS-ResNet} & 238K & 239K & 97.89 & 97.31 \\ \hline
			Transformer~\cite{KWS-Transformer} & 232K & 234K & 96.21 & 97.14 \\
			\hline \hline
		\end{tabular}
	}
\end{table}

\noindent \textbf{Networks and Centralized Training}:
We investigate various network architectures and moderately modify them to keep nearly the same number of parameters. We use DSCNN~\cite{KWS-DSCNN} with 172 channels, MHAttRNN~\cite{KWS-MHAttRNN,KWS-Transformer} with 4 heads and 80 hidden neurons, ResNet~\cite{KWS-ResNet} with 15 layers and 45 channels in each basic block, Transformer~\cite{KWS-Transformer} with 4 layers and a model dimension of 96. We first use these networks for centralized training. For DSCNN, MHAttRNN and ResNet, we utilize SGD optimizer with momentum 0.9, and we vary the learning rate in $\{0.1, 0.05, 0.03, 0.01\}$ and select the best result. For Transformer, we utilize AdamW optimizer and vary learning rate in $\{0.005, 0.002, 0.0008\}$. We set batch size as 128. The number of network parameters and the accuracies on test data are shown in Table~\ref{tab:net}. We do not obtain SOTA results via Transformer because we only use 4 layers with 0.23M parameters while~\cite{KWS-Transformer} uses a network with up to 5.4M parameters.

\begin{table}
	\centering
	\caption{\small Comparisons on FA and FR rate. The lower the better.}
	\label{tab:fa-fr}
	{
		\begin{tabular}{l|c|c|c}
			\hline \hline
			& FedAvg~\cite{FedAvg,FedKWS-Spec} & FedOpt~\cite{FedOpt,FedKWS-Ada} & FedKWS-UI  \\
			\hline \hline
			FA & 0.27 & 0.32 & {\bf 0.23} \\ \hline
			FR & 5.03 & 3.19 & {\bf 2.78} \\
			\hline \hline
		\end{tabular}
	}
\end{table}

\begin{figure}[htb]
	\centering
	\centerline{\includegraphics[width=\linewidth]{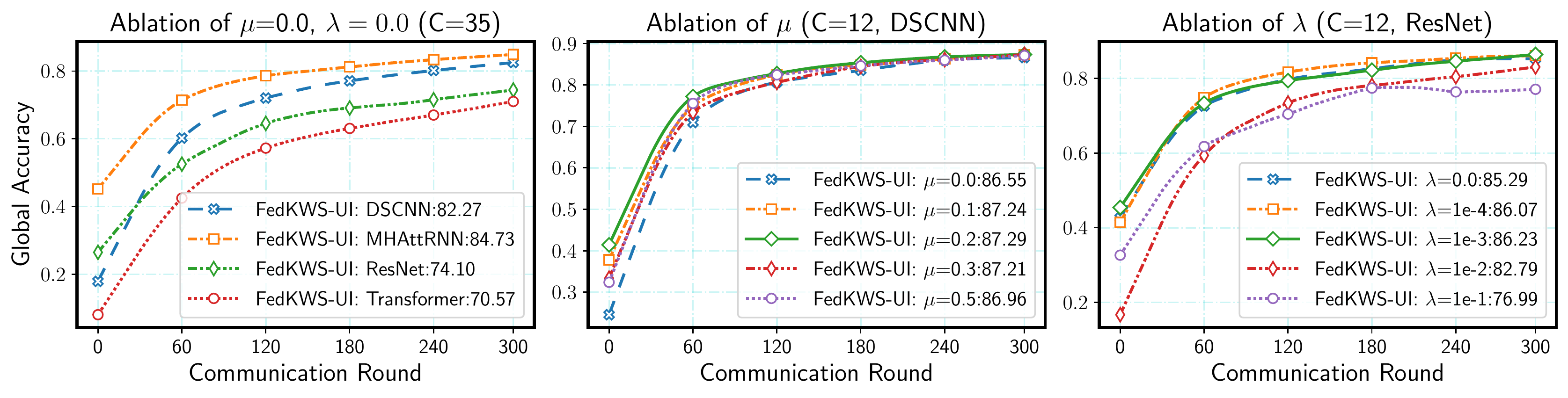}}
	\caption{Ablation studies of only using ALT (\textbf{left}) and the hyper-parameters in ALO ($\mu$ (\textbf{middle}), and $\lambda$ (\textbf{right})).}
	\label{fig:ablation}
\end{figure}

\noindent \textbf{FedKWS}:
For FedKWS, we split the training data in Google Speech Commands via the provided user IDs. The test data is still used to evaluate the generalization ability of the aggregated model. We plot the statistics of clients' number of samples and class distribution entropy at the right of Figure~\ref{fig:data}. We compare FedKWS-UI with FedAvg~\cite{FedAvg} (used in~\cite{FedKWS-Spec}), FedProx~\cite{FedProx}, FedMMD~\cite{FedMMD}, FedOpt~\cite{FedOpt} (used in~\cite{FedKWS-Ada}). For all methods, we use a batch size of 32, local training steps $E=50$ and run 300 rounds. We also vary the learning rate as aforementioned and take the best one for comparison. Additionally, for FedProx and FedMMD, we vary the regularization coefficient in $\{0.0001, 0.001, 0.01\}$. For FedOpt, we vary the global optimizer in \{SGD, Adam\} and the global learning rate in $\{1.0, 0.1\}$ and $\{0.001,0.0001\}$, respectively. For FedKWS-UI, we use $r_0=3.5,5.0$ for $C=12,35$, and show $r_0 * \overline{r}_k$ values of all clients via the shades of color at the right of Figure~\ref{fig:data}. The max and min values are shown at the color bar, and the top-right points (clients) tend to have larger $r_0 * \overline{r}_k$. We utilize $\mu=0.2$ and $\lambda=0.001$ in FedKWS-UI (Eq.~\eqref{eq:ls}, Eq.~\eqref{eq:loss}).


We record the accuracy on the global test set every 3 rounds and plot the convergence curves in Figure~\ref{fig:compare}. First, we can clearly observe that the decentralized performances drop a lot compared with centralized training. For example, all of the compared methods could only obtain accuracy as high as 86.39 on task 12, far away from the centralized training (97.89). Then, comparing the network architectures, we could find that MHAttRNN tends to obtain higher performances while Transformer performs worst. We guess that MHAttRNN could be more robust to the random time-shift because it directly computes the sequential information, and the attention mechanism could precisely capture the important signals. Furthermore, FedProx and FedMMD add regularization during local training procedures on the parameter and intermediate features, which perform not so well. Overall, FedKWS-UI could lead to better results on all of these architectures, particularly on ResNet and Transformer, verifying the versatility of our methods. Significantly, FedKWS-UI surpasses all compared methods by a large margin on task 12 with DSCNN, ResNet, and Transformer. For example, FedKWS-UI could boost the Transformer performances on task 12 from 72.71 to 79.06. We also evaluate the false accept (FA) and false reject (FR) rate as done in~\cite{FedKWS-Ada,FedKWS-Spec}. In task 12, we take the 10 words as positive classes and average their FA rates, while the silence and unknown as negative classes. We do not adjust the prediction confidence threshold to control the FA and directly report FA and FR with the predictions. We use DSCNN and calculate the FA and FR rate of the final aggregated model. We show the results in Table~\ref{tab:fa-fr}. We find that FedKWS-UI could obtain fewer false alarms/rejections.

\noindent \textbf{Ablation Studies}:
We investigate the effects of components in FedKWS-UI. First, we set $\mu=0$ and $\lambda=0$ to omit the part of adversarial learning against overfitted models (ALO) and only use adaptive local training (ALT). We record the results using four networks on task 35 at the left of Figure~\ref{fig:ablation}. We find that only using adaptive local training could still perform well on MHAttRNN and Transformer, while it works worse on DSCNN and especially on ResNet. Hence, it is still necessary to utilize the adversarial learning to improve performances further. Then, we vary $\mu \in \{0.0, 0.1, 0.2, 0.3, 0.5\}$ and $\lambda \in \{0.0, 0.0001, 0.001, 0.01, 0.1\}$ correspondingly, studying the effects of label smoothing and adversarial loss in ALO. We investigate the task $C=12$ with DSCNN and ResNet. The results are shown at the middle and right of Figure~\ref{fig:ablation}. Utilizing label smoothing could almost lead to better performances, and setting $\mu$ around 0.2 is a better choice. Similarly, $\lambda=0.001$ is recommended for the proposed adversarial loss, and a larger $\lambda$ (e.g., 0.1) could be harmful.

\begin{figure}[htb]
	\centering
	\centerline{\includegraphics[width=\linewidth]{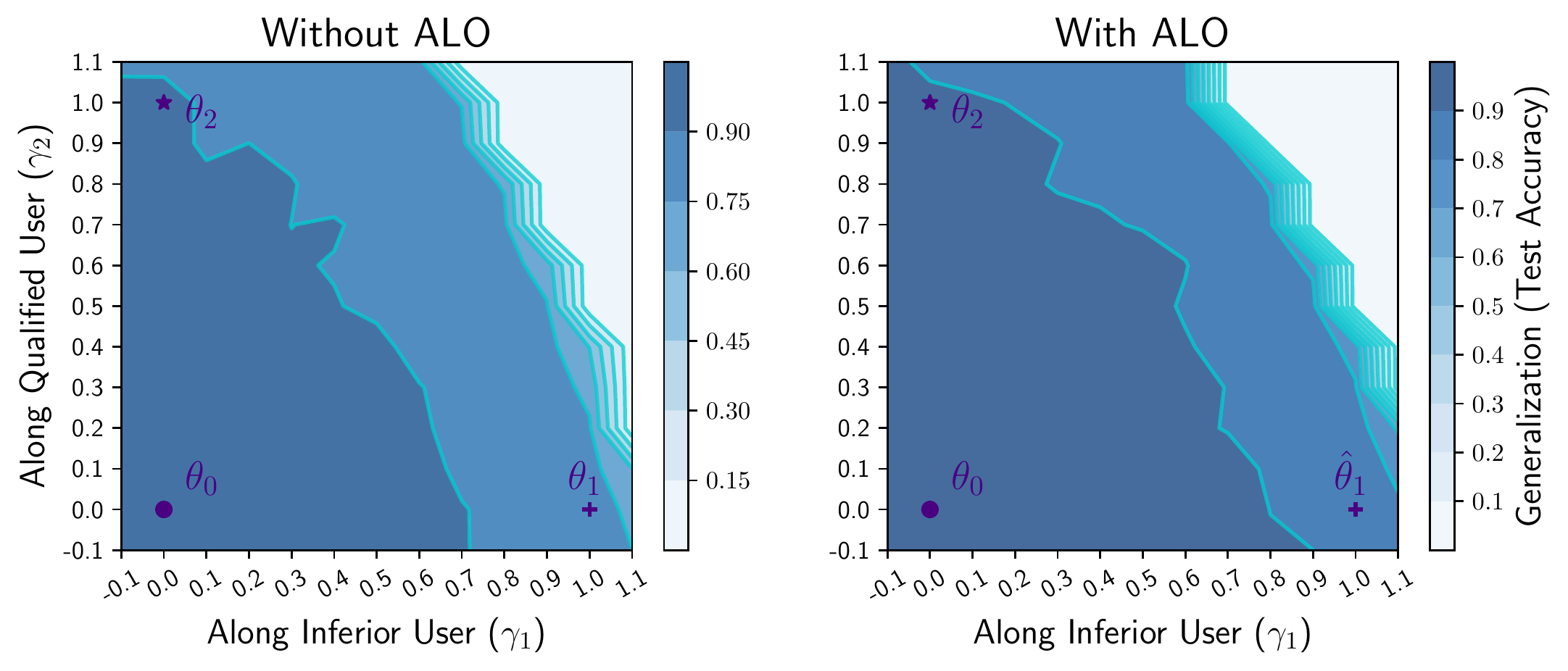}}
	\caption{Visualization of the plausibility and advantage of the proposed strategies.}
	\label{fig:scape}
\end{figure}

\noindent \textbf{Visualization Analysis}:
We then present some visualization results on task 35 to further show the plausibility and advantage of the proposed strategies. Specifically, we first train a well-performed KWS model ($\theta_0$) on the centralized training set (test accuracy up to 93.0\%). Then, we respectively update $\theta_0$ on an inferior and qualified user's data for $20$ epochs. The inferior user owns only 100 samples and the classes are imbalanced (i.e., the bottom-left user shown in the right part of Figure~\ref{fig:data}), while the qualified user owns about 250 samples and the classes are more balanced (i.e., the top-right user shown in the right part of Figure~\ref{fig:data}). The updated models are denoted as $\theta_1$ and $\theta_2$. Then, we plot the performance landscape of the interpolation $\theta_0 + \gamma_1 (\theta_1 - \theta_0) + \gamma_2 (\theta_2 - \theta_0)$ within the grid space where $\gamma_1 \in [-0.1, 1.1], \gamma_2 \in [-0.1, 1.1]$. The left part of Figure~\ref{fig:scape} shows that updating $\theta_0$ on the qualified user's data keeps the generalization ability of the global model while the result on the inferior user's data becomes worse. Hence, it is rational that our proposed ALT encourages qualified users to contribute more to FedKWS. Additionally, for the inferior user, we utilize the proposed ALO to train another model $\hat{\theta}_1$ against the overfitted $\theta_1$, and $\hat{\theta}_1$ performs better as shown on the right of Figure~\ref{fig:scape}. The interpolation landscape along the inferior user's data becomes smoother with ALO, benefiting the model aggregation procedure in FL. This verifies the advantage of the proposed ALO. Overall, FedKWS-UI could enhance the generalization ability of the federated model even with few or skewed samples.

\section{Conclusion}
We investigate popular networks for FedKWS, where the data heterogeneity leads to significant performance degradation compared with centralized training. We propose to learn user-invariant information via adversarial learning against overfitted local models and a computation re-allocation strategy named adaptive local training. These two strategies could avoid overfitting user-specific information during local training and facilitate model aggregation. Experimental results verify the superiorities of our proposed FedKWS-UI. Future works will extend this work to streaming KWS~\cite{KWS-MHAttRNN} and utilize differential privacy~\cite{DeepDP} to satisfy stricter privacy requirements.


\newpage
\bibliographystyle{IEEEtran}
\bibliography{refs}

\end{document}